\documentclass{article}

 \usepackage[dblblindworkshop, final]{neurips_2025}
\workshoptitle{Advances in Representation Learning for Earth Observation}

\usepackage[utf8]{inputenc} 
\usepackage[T1]{fontenc}    
\usepackage{hyperref}       
\usepackage{url}            
\usepackage{booktabs}       
\usepackage{amsfonts}       
\usepackage{nicefrac}       
\usepackage{microtype}      
\usepackage{xcolor}         
\usepackage{graphicx}
\usepackage{multirow}
\usepackage{makecell}
\usepackage{wrapfig}
\usepackage{graphicx}
\usepackage{subcaption}

\title{Gold Exploration using Representations \\from a Multispectral Autoencoder}

\author{%
  Argyro \textsc{Tsantalidou} \\
  Technology Innovation Institute\\
  Abu Dhabi, UAE \\
  \texttt{Argyro.Tsantalidou@tii.ae} \\
  \And
  Konstantinos \textsc{Dogeas}\\
  Technology Innovation Institute\\
  Abu Dhabi, UAE \\
  \texttt{Konstantinos.Dogeas@tii.ae} \\
  \And
  Eleftheria \textsc{Tetoula Tsonga} \\
  Institute of Communication \\and Computer Systems\\
  Athens, Greece \\
  \texttt{Tetoula.Tsonga@iccs.gr} \\
  \And
  Elisavet \textsc{Parselia} \\
  Geonova\\
  Athens, Greece \\
  \texttt{Elisavet@geonova.ai} \\
  \And
  Georgios \textsc{Tsimiklis} \\
  Institute of Communication \\and Computer Systems\\
  Athens, Greece \\
  \texttt{Georgios.Tsimiklis@iccs.gr} \\
  \And
  George \textsc{Arvanitakis} \\
  Technology Innovation Institute\\
  Abu Dhabi, UAE \\
  \texttt{George.Arvanitakis@tii.ae} \\
}

\begin{document}

\maketitle

\begin{abstract}
Satellite imagery is employed for large-scale prospectivity mapping due to the high cost and (typically) limited availability of on-site mineral exploration data. In this work, we present a proof-of-concept framework that leverages  generative representations learned from multispectral Sentinel-2 imagery to identify gold-bearing regions from space. An autoencoder foundation model, called Isometric, which is pretrained on the large-scale FalconSpace-S2 v1.0 dataset, produces information-dense spectral–spatial representations that serve as inputs to a lightweight XGBoost classifier. We compare this representation-based approach with a raw spectral input baseline using a dataset of 63 Sentinel-2 images from known gold and non-gold locations. The proposed method improves patch-level accuracy from 0.51 to 0.68 and image-level accuracy from 0.55 to 0.73, demonstrating that generative embeddings capture transferable mineralogical patterns even with limited labeled data. These results highlight the potential of foundation-model representations to make mineral exploration more efficient, scalable, and globally applicable.
\end{abstract}
\section{Introduction}
%
Mineral exploration is the process of identifying naturally occurring mineral deposits.
As supply chains increasingly rely on these resources, efficient exploration is essential to ensure their stability \cite{eu2024rawmaterials}. 
Assessing their economic viability for extraction and industrial use remains a crucial problem. 
Traditionally, mineral exploration has relied on direct field observations to infer mineral composition; and while highly accurate, these methods are time-consuming, costly, and often impractical for remote or inaccessible regions. 
Nevertheless, the availability of high-resolution satellite imagery from missions such as Sentinel-2 and EnMAP provides complementary data that can reveal spectral and spatial patterns that reflect surface mineralogy that can potentially enable broader, more efficient exploration.
 
Machine Learning techniques are well-suited to leverage image data, as they can learn complex relationships between spectral–spatial patterns and potentially underlying mineral distributions. 
To this extent, several supervised models, including Support Vector Machines, Neural Networks, and Transformers, have been used to predict mineralogical composition directly from imagery \cite{han2023survey}. 
However, these approaches require large amounts of labeled data, which are costly to obtain and often limit generalization across different sensors and regions. 
To address this limitation, recent studies in generative and self-supervised learning, such as Masked Autoencoders and diffusion-based models, demonstrate the ability to learn meaningful spectral–spatial structures from unlabeled data \cite{liu2024diffusionsat}. 
The representations produced through this process capture and organize the essential information of the input data. 
In satellite imagery, they capture spatial patterns and spectral behavior~\cite{wang2024spectralgpt}. 

\paragraph{Our contribution}
We create a new dataset for gold exploration. 
We leverage information from \url{Mindat.org} to identify locations of gold mines.
Sentinel-2 images are downloaded for these locations with the addition of randomly sampled locations from the Earth.

We propose a gold exploration framework using Sentinel-2 data, where spectral–spatial representations from a generative model are fed into an XGBoost classifier. 
We compare the performance of this \textbf{Representations}-based method to the performance of the same XGBoost classifier that uses \textbf{Raw Input} (i.e. the $12$ multispectral bands of the Sentinel-2 images).
The \emph{Representations} approach outperforms the \emph{Raw Input} approach, improving patch-level accuracy from $0.52$ to $0.68$ and image-level accuracy from $0.55$ to $0.73$, showing the value of representation-based features for mineral exploration.
We also tested the performance of the current state of the art in multispectral representation learning (SpectralGPT), and we find that the Isometric model achieves better performance.

\section{Model}
\subsection{Generative AI Model}
\textbf{FalconSpace-S2v1.0 Dataset}: To train our foundation model, we constructed the FalconSpace-S2v1.0 dataset comprising $1.156.800$ multispectral Sentinel-2 images, each with a size of $128 \times 128$ pixels with $12$ spectral bands. To ensure a comprehensive representation of Earth's surface conditions, we implemented three sampling strategies: (1) Temporal diversity—acquisition dates selected uniformly between 01/01/2020 and 30/06/2025, capturing seasonal and environmental variations throughout the year; (2) Spatial diversity—uniform sampling across latitudes and longitudes worldwide; (3) Land cover balancing—weighted sampling according to land use/land cover (LULC) classes, giving extra weight to less representative classes and built-up areas. Finally, we ensure that all images have cloud coverage below $30\%$. 

\textbf{Autoencoder Architecture:} The generative AI model used in this study follows an autoencoder architecture pretrained on the large-scale FalconSpace-S2 v1.0 dataset. 
Its design is based on the classical transformer-based Masked Autoencoders (MAEs,~\cite{he2022masked}) that learn high-fidelity spectral–spatial representations from unlabeled data. 
The encoder is specifically adapted for multispectral imagery, following a design similar to SpectralGPT, introduced in~\cite{wang2024spectralgpt}, where the spectral bands are grouped into $8 \times 8 \times 3$ patch tokens that preserve both spatial context and spectral continuity. 
In contrast to SpectralGPT, our design employs a deep decoder, with the main focus being the reconstruction quality of the full spectral cube. 
This gives a boost in the quality of the representations.
During pretraining, random parts of the input are masked, forcing the encoder to infer missing content and capture long-range dependencies. 
In contrast to the commonly high masking ratio ($90\%$) followed in the literature of vision transformers, we apply a lower masking ratio ($40\%$) that helps the performance of the model during inference.
Once trained, the encoder is \textbf{frozen} and serves as a feature extractor: its latent embeddings are later leveraged as fixed, reusable inputs in our mineral exploration pipeline to distinguish gold-bearing from non gold-bearing regions using a lightweight classifier.
We apply the same strategy in both the Isometric and SpectralGPT models.

\paragraph{Reconstruction Performance:} In order to assess the reconstruction performance of the proposed autoencoder, we use the following widely used metrics: Mean Squared Error (MSE) measures pixel-wise differences between original and reconstructed images; 
Peak Signal-to-Noise Ratio (PSNR) expresses reconstruction fidelity in decibels, with higher values indicating better quality; 
Structural Similarity Index (SSIM) assesses perceptual similarity based on luminance, contrast, and structure; 
Spectral Angle Mapper (SAM) quantifies spectral distortion by measuring the angle between original and reconstructed spectra; 
and Relative Global Dimensional Error (ERGAS) evaluates overall radiometric consistency across all spectral bands. 
The following table shows the mean $\pm$ standard deviation for each metric of our model compared to the current state of the art, i.e. SpectralGPT(~\cite{wang2024spectralgpt}). 
The arrows $\downarrow$ and $\uparrow$ denote \textit{the smaller the better} and \textit{the higher the better}, respectively.

\begin{table}[ht]
\centering
\small
\caption{Reconstruction performance metrics}
\resizebox{1\textwidth}{!}{%
    \begin{tabular}{l||ccccc}
    \hline
    \textbf{Method} & \textbf{MSE} $\downarrow$ & \textbf{SAM} $\downarrow$ & \textbf{ERGAS} $\downarrow$ & \textbf{PSNR} $\uparrow$ & \textbf{SSIM} $\uparrow$ \\
    \hline
    SpectralGPT  & 0.062 $\pm$ 0.037 & 0.217 $\pm$ 0.248 & 17.993 $\pm$ 49.820 & 21.890 $\pm$ 3.971 & 0.612 $\pm$ 0.0918 \\
    Isometric    & \textbf{0.006 $\pm$ 0.005} & \textbf{0.093 $\pm$ 0.261} & \textbf{8.359 $\pm$ 35.024} & \textbf{32.784 $\pm$ 3.015} & \textbf{0.942 $\pm$ 0.056} \\
    \hline
    \end{tabular}
    \label{tab:reconstructions}
}
\end{table}

\subsection{Mineral Exploration Model}
\textbf{Gold Exploration Dataset}: To evaluate our representations on mineral exploration, we collected $63$ Sentinel-2 images from June to August 2023: $33$ images are taken from known gold deposit locations (scraped from \url{Mindat.org}) and $30$ images are taken from random land locations distributed globally. 
Each $128 \times 128 \times 12$ image was divided into $1,024$ patches of $8 \times 8 \times 3$  pixels. 
This process, applied to our $63$ images, produced $33,792$ patches from gold locations and $30,720$ patches from non-gold locations. 
To create training and testing sets, we performed a $80/20$ cross validation split at the image level, resulting in $52,224$ patches (from $51$ images) for training and $12,288$ patches (from $12$ images) for testing in each split. 


We train an XGBoost patch-level classifier that assigns each patch a probability of belonging to a gold or non-gold location. 
During inference, we classify all $1,024$ patches of an image as gold or non-gold using the XGBoost.
We then aggregate these patch-level predictions through majority voting to reach a final image-level decision: gold or non-gold \emph{location}. 

This pipeline enables us to directly compare the discriminative power of learned representations versus raw multispectral input for mineral exploration.

\section{Results}
To evaluate the effectiveness of the learned representations compared to the raw data 
and the state of the art SpectralGPT, we evaluated the performance at both the patch and image levels. 
\begin{wrapfigure}{r}{0.55\textwidth} 
  \centering
  \vspace{-15pt} 
  \includegraphics[width=0.95\linewidth]{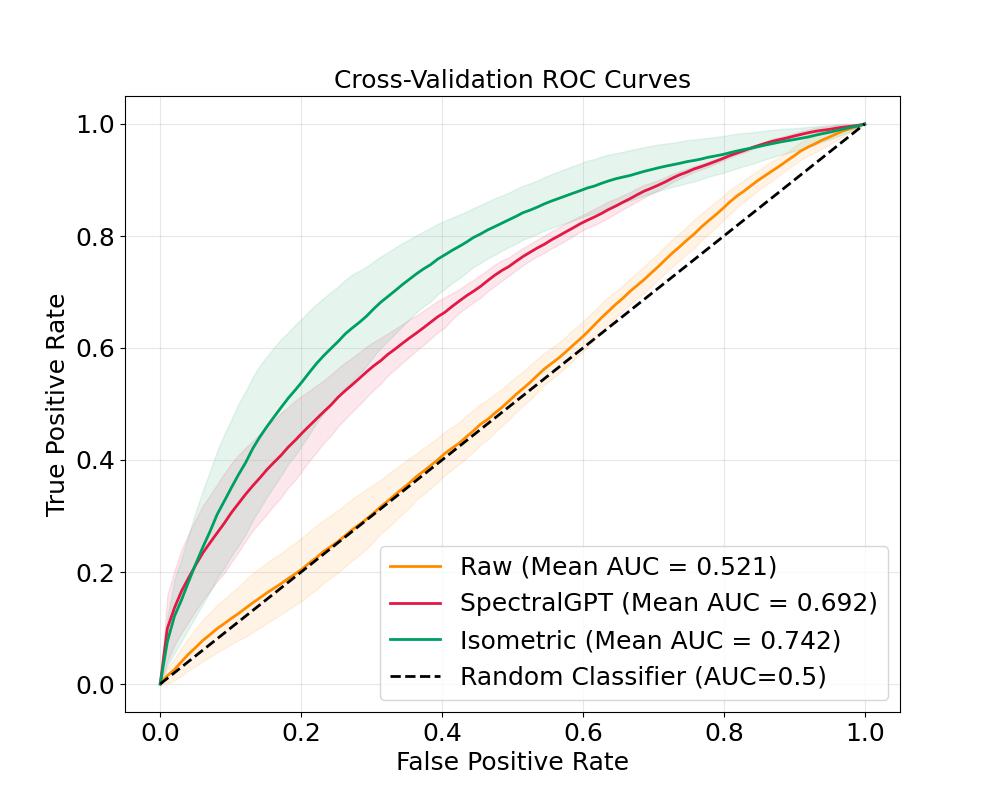}
  \caption{\footnotesize ROC curves illustrating the classification performance of the Raw, Spectral, and Isometric approaches.}
  \label{fig:auc}
  \vspace{-15pt} 
\end{wrapfigure}
At the patch level (see Table \ref{tab:patch-level}), our representation-based approach achieved an accuracy of $0.68$, compared to $0.51$ for the raw baseline, achieving a considerable improvement.
Our method is consistently better across the Precision, Recall, and F1-score metrics.



The patch-level Receiver Operating Characteristic-Area Under the Curve (ROC-AUC) also improved from $0.52$ to $0.74$, indicating better discriminative capabilities improved by $42\%$, as depicted in Figure \ref{fig:auc}. 

At the image level (see Table~\ref{tab:image-level}), the accuracy increased from $0.55$ to $0.73$, confirming that the embeddings trained via generative reconstruction capture relevant spectral–spatial patterns associated with mineralization zones. 
These findings suggest that frozen generative encoders can effectively serve as universal feature extractors for resource exploration, even with limited training data.
Additionally, the ROC curves demonstrate that our encoder outperforms the state of the art multispectral representation model, achieving an improvement of $7.2\%$ (see Figure~\ref{fig:auc}).

\textit{Note: Given the limited amount of available data, all experiments were conducted using a $5$-fold
cross-validation scheme, with multiple random initializations of the train–test splits.}

\begin{table}[ht]
\centering
\small 
\caption{Patch-level metrics (mean across folds)}
\begin{tabular}{lccccc}
\toprule
\textbf{Approach} & \textbf{Accuracy} & \textbf{Precision} & \textbf{Recall} & \textbf{F1 score} \\
\midrule
Raw & 0.517 $\pm$ 0.010 & 0.513 $\pm$ 0.012 & 0.512 $\pm$ 0.012 & 0.508 $\pm$ 0.012 \\
SpectralGPT & 0.630 $\pm$ 0.011 & 0.642 $\pm$ 0.012 & 0.628 $\pm$ 0.014 & 0.618 $\pm$ 0.015 \\
Isometric & \textbf{0.681 $\pm$ 0.043} & \textbf{0.692 $\pm$ 0.042} & \textbf{0.680 $\pm$ 0.041} & \textbf{0.674 $\pm$ 0.044} \\
\bottomrule
\end{tabular}
\label{tab:patch-level}
\end{table}
%
%
\begin{table}[ht]
\centering
\small
\caption{Image-level metrics (mean across folds)}
\begin{tabular}{lcccc}
\toprule
\textbf{Approach} & \textbf{Accuracy} & \textbf{Precision} & \textbf{Recall} & \textbf{F1 score} \\
\midrule
Raw & 0.554 $\pm$ 0.101 & 0.551 $\pm$ 0.205 & 0.541 $\pm$ 0.098 & 0.488 $\pm$ 0.130 \\
SpectralGPT & 0.635 $\pm$ 0.039 & 0.683 $\pm$ 0.064 & 0.626 $\pm$ 0.049 & 0.600 $\pm$ 0.068 \\
Isometric & \textbf{0.733 $\pm$ 0.130} & \textbf{0.752 $\pm$ 0.141} & \textbf{0.733 $\pm$ 0.129} & \textbf{0.729 $\pm$ 0.131} \\
\bottomrule
\end{tabular}
\label{tab:image-level}
\end{table}

\begin{figure}[h]
    \centering

    \begin{subfigure}[t]{0.49\linewidth}
        \centering
        \includegraphics[width=.32\linewidth]{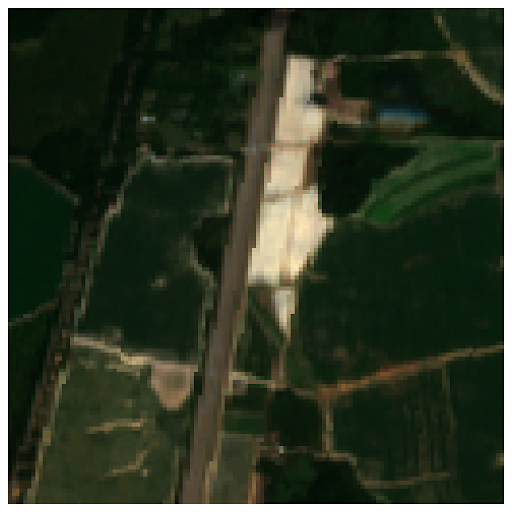}
        \includegraphics[width=.32\linewidth]{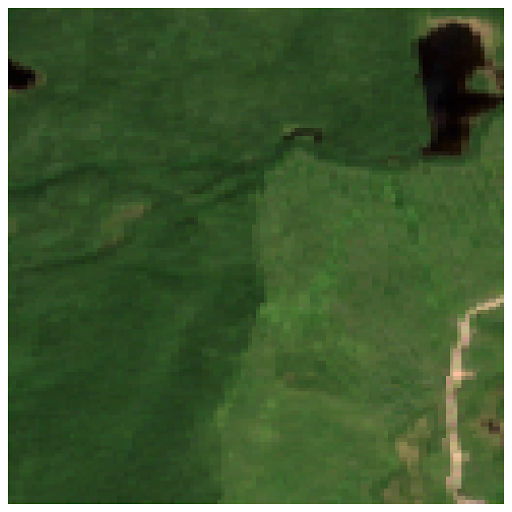}
        \includegraphics[width=.32\linewidth]{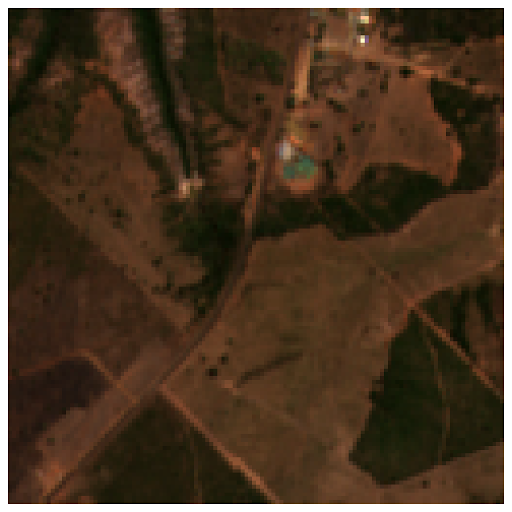}
        \caption{Non Gold samples}
        \label{fig:results_mineral_left}
    \end{subfigure}
    \hfill
    \begin{subfigure}[t]{0.49\linewidth}
        \centering
        \includegraphics[width=.32\linewidth]{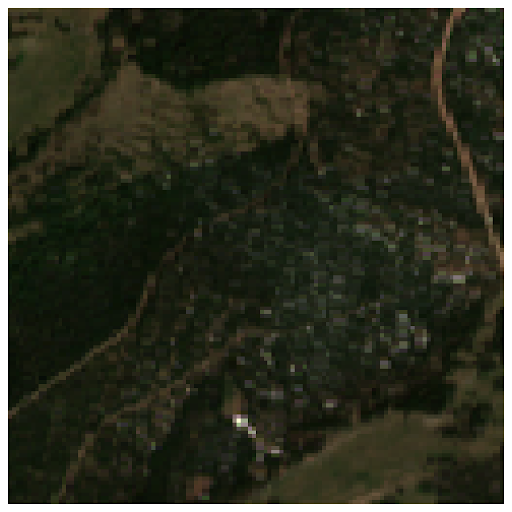}
        \includegraphics[width=.32\linewidth]{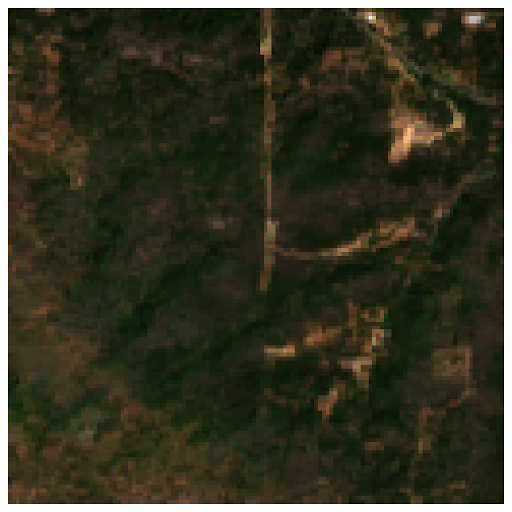}
        \includegraphics[width=.32\linewidth]{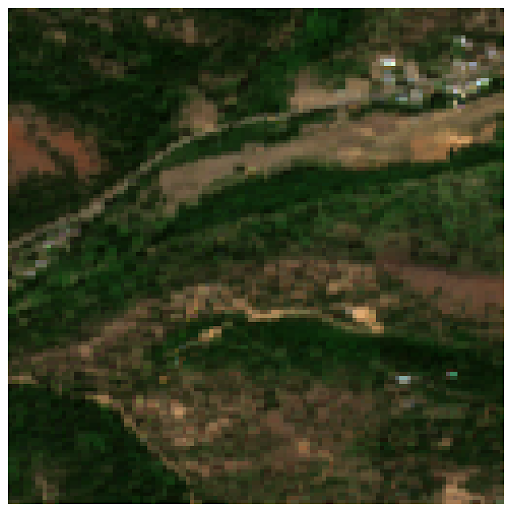}
        \caption{Gold samples}
        \label{fig:results_mineral_right}
    \end{subfigure}

    \caption{Gold vs Non-Gold. The three images of subfigure~\ref{fig:results_mineral_left} do not contain gold, while the three images of subfigure~\ref{fig:results_mineral_right} do. 
    Using the representations of the Isometric model (our approach), XGBoost classifies correctly all six images, while using the raw input, XGBoost mislabel two out of three samples that contain gold (only the middle image of subfigure~\ref{fig:results_mineral_right} is correctly predicted using raw data).}
    \label{fig:Results_mineral}
\end{figure}

In Figure \ref{fig:Results_mineral}, we show a few images from our dataset using only the visible RGB bands ($3$ of the $12$ Sentinel-2 channels). 
Notably, the scenes appear practically indistinguishable to the human eye, regardless of whether they correspond to gold-bearing locations or not.

%
\section{Conclusion}
We presented a proof-of-concept study demonstrating how generative representations can enhance mineral prospectivity mapping from space. By coupling Sentinel-2 imagery with an autoencoder architecture and a lightweight XGBoost classifier, we achieved notable performance gains over raw spectral inputs. Despite using a dataset of $63$ images, the results show that generative embeddings capture and can generalize across unseen regions. The frozen encoder can be reused as a fixed feature extractor for different minerals or locations, requiring only simple task-specific classifiers trained with limited labeled data. This makes satellite-based mineral prospecting more efficient, scalable, and accessible.
Future work will apply this framework to larger mineral datasets, integrate SAR and hyperspectral data, and include multi-temporal analysis to better capture seasonal and surface variation in alteration zones.



\bibliographystyle{plainnat}
\bibliography{main.bib}



\end{document}